\pdfoutput=1

\documentclass[11pt]{article}

\usepackage{authblk}
\usepackage[]{acl}

\usepackage{times}
\usepackage{latexsym}

\usepackage[T1]{fontenc}

\usepackage[utf8]{inputenc}

\usepackage{microtype}

\usepackage{inconsolata}

\usepackage{enumitem}
\usepackage{tabularx}

\usepackage{tcolorbox}
\tcbuselibrary{listings,breakable,skins}

\usepackage{multirow}
\usepackage{graphicx}
\usepackage[normalem]{ulem}
\useunder{\uline}{\ul}{}
\usepackage{booktabs}

\usepackage{textcomp}

\usepackage{caption}
\usepackage{subcaption}
\newcounter{subfig} 

\newcommand{\fic}{\texttt{FiC}}
\newcommand{\frev}{\textsc{FuseReviews}\includegraphics[height=1em]{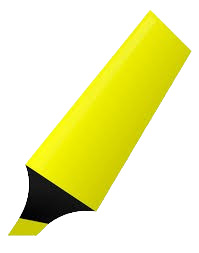}}

\newcommand{\ran}[1]{{\textcolor{orange}{[ran] #1}}}
\newcommand{\os}[1]{{\textcolor{blue}{[os] #1}}}
\newcommand{\avivs}[1]{{\textcolor{red}{[aviv] #1}}}
\newcommand{\oscr}[1]{{\textcolor{blue}{[os] #1}}}
\renewcommand{\ran}[1]{}
\renewcommand{\os}[1]{}
\renewcommand{\avivs}[1]{}
\renewcommand{\oscr}[1]{}

\title{Multi-Review Fusion-in-Context}

\author[ \; 1,2]{\bf Aviv Slobodkin\thanks{\;\; Work was done during an internship at Amazon.}}
\author[2]{\bf Ori Shapira}
\author[2]{\bf Ran Levy}
\author[1]{\bf Ido Dagan}

{
\makeatletter
\renewcommand\AB@affilsepx{~~~~~~ \protect\Affilfont} \makeatother

\affil[1]{Bar-Ilan University}
\affil[2]{Amazon}
}
\affil[  ]{} 
\affil[  ]{\tt lovodkin93@gmail.com}
\affil[  ]{\tt \{orishap, ranlevy\}@amazon.com}
\affil[  ]{\tt dagan@cs.biu.ac.il}

\begin{document}
\maketitle

\begin{abstract}

Grounded text generation, encompassing tasks such as long-form question-answering and summarization, necessitates both content selection and content consolidation. Current end-to-end methods are difficult to control and interpret due to their opaqueness.
Accordingly, recent works have proposed a modular approach, with separate components for each step. Specifically, we focus on the second subtask, of generating coherent text given pre-selected content in a multi-document setting. Concretely, we formalize \textit{Fusion-in-Context} (\fic{}) as a standalone task, whose input consists of source texts with highlighted spans of targeted content. A model then needs to generate a coherent passage that includes all and only the target information.
Our work includes the development of a curated dataset of 1000 instances in the reviews domain, alongside a novel evaluation framework for assessing the faithfulness and coverage of highlights, which strongly correlate to human judgment. Several baseline models exhibit promising outcomes and provide insightful analyses.
This study lays the groundwork for further exploration of modular text generation in the multi-document setting, offering potential improvements in the quality and reliability of generated content.\footnote{Our benchmark, \frev{}, including the dataset, evaluation framework, and designated leaderboard, can be found at \url{https://fusereviews.github.io/}.}
\end{abstract}

\begin{figure*}[h!]
\centering
    \includegraphics[width=0.9\linewidth]{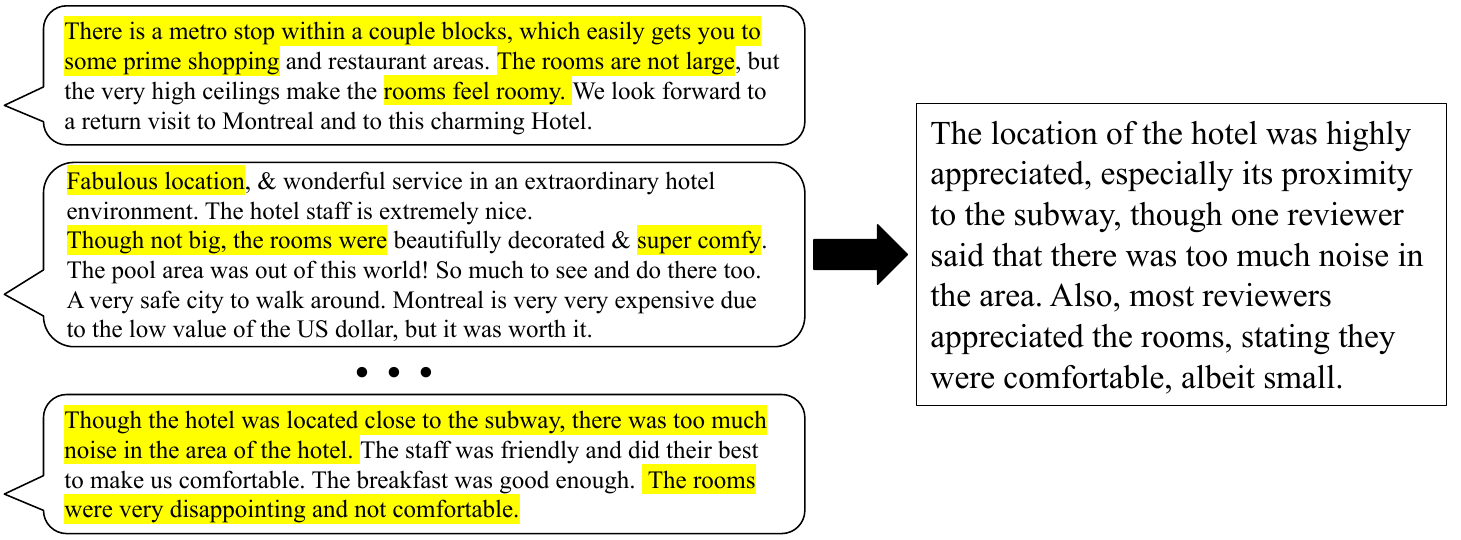}
    \caption{An example of an input, consisting of multiple reviews with highlights (left), and the generated text fusing the highlighted content while preserving coherence and non-redundancy (right). Such highlights in realistic use cases may be produced by different content-selection strategies.}
    \label{fig:example_task}
\end{figure*}

\section{Introduction}\label{sec:introduction}
Grounded text generation is the task of producing a passage from source texts, where the output is anchored around use-case-dependent spans within the source texts.
Although such a task involves two distinct subtasks -- identifying relevant spans and fusing them -- it is commonly handled in an end-to-end approach, recently by using Large Language Models (LLMs) \citep{shuster_blenderbot_2022, su2022read, zhang2023benchmarking}.
While effective, this approach often lacks flexibility and control over the generation process, given its opaque nature.



Addressing this, \citet{slobodkin-etal-2022-controlled} recently advocated splitting grounded generation tasks into their two subtasks, and particularly focused on the fusion step. They introduced \textit{Controlled Text Reduction} (CTR), a task where pre-selected spans in a source document (`highlights') are fused into a coherent text that exclusively covers the spans. 
This approach enhances control and modularity in text generation, enabling a single CTR model to work with various content selection strategies and user preferences, applicable in different contexts like summarization or long-form question-answering.
It could also support human-in-the-loop scenarios for tailored outputs based on user preferences, as explored in \citet{slobodkin2023summhelper}.
Further, the direct access to the highlights that contribute to the output facilitates attributed generation \citep{bohnet_attributed_2023, gao-etal-2023-rarr, gao_enabling_2023}, where models can cite source spans for generated text. 

Despite its benefits, CTR's focus on single-input scenarios limits its applicability to the broader, and more complex, multi-document setting. In this paper, we bridge this gap and extend the task to the multi-document setting. 
For that, we introduce the task of \textit{Fusion-in-Context} (\fic{}), a generalized version of the CTR task, which processes multiple documents with pre-selected highlights, and aims to fuse them into a coherent, non-redundant text covering all and only the highlighted content, as demonstrated in \autoref{fig:example_task}. 
In addition to the challenges of the single-input CTR task, including coreference resolution and proper discourse for coherence, the multi-document setting also requires handling repetitive, and sometimes conflicting information \citep{ma2020multi}. Specifically, our work focuses on the business reviews domain, where contradicting opinions are more prevalent than in other more fact-oriented domains, such as news.

\ran{Maybe here say something like: In addition, the reviews domain exemplifies the usefulness of breaking the end-to-end task into marking highlights and fusing them. For example, users can define their own criteria for span highlighting, e.g., highlight positives statements about aspects X,Y,Z taken from reviews from the last year. Then the same fuser, can generate on-the-fly a coherent summary of the highlights, regardless of what highlights were chosen.}

To promote research on \fic{}, we start by formally defining the task (\S\ref{sec:task_definition}). We then introduce a dataset (\S\ref{sec:crowdsourcing}), carefully constructed via controlled crowdsourcing \citep{roit-etal-2020-controlled}. Each of its 1000 instances comprises a set of inputs with highlights, and a corresponding fused text. The dataset is created through an efficient procedure, adapted from \citet{slobodkin-etal-2022-controlled}, leveraging existing multi-document summarization datasets, specifically in the business reviews domain.
We also develop an evaluation framework (\S\ref{sec:evaluation_framework}) that assesses outputs' faithfulness and coverage of highlights. The dataset and evaluation framework are released as a benchmark, called \frev{}. We explore various baseline models on the benchmark and report their performance (\S\ref{sec:experiments}).
Our findings reveal that while these models show promising results, there is still room for further improvement in future research.

\ran{I think we should provide a few example use cases for different selections that make sense: e.g. user asks to highlight all review excerpts mentioning x,y,z aspects with postivie sentiment and the same with negative sentiment and then fuse each}
\ran{futrue work can also discuss different fusers, e.g. fusers that generate graphs and text}

\os{Possibility to add for motivation: in cases where there are many reviews (cite MMDS paper) it is worth allowing a user to filter reviews such as by aspect and recency. Then the fuser can be used independently of the selected reviews/content.}
\section{Background}\label{sec:background}
Grounded text generation, an area focusing on generating text from source documents, requires identifying relevant task-specific details within the inputs, such as salient content for summarization, as well as their coherent fusion.
This field includes tasks like long-form question-answering \citep{fan2019eli5, stelmakh_asqa_2023}, summarization \citep{nallapati2016abstractive, nallapati2016classify, shapira2020massive, brazinskas-etal-2020-unsupervised, zhao2022mrs}, and dialogue systems \citep{yan2017building, xu2019end, thoppilan_lamda_2022}, with most related datasets aimed at end-to-end training \citep{fan2019eli5, bravzinskas2020few, liu2021durecdial, iso2022comparative}.


Despite the prevalence of end-to-end systems, there has been a growing trend towards decomposed pipeline approaches, particularly in summarization, with several recent studies focusing on content selection \citep{gehrmann-etal-2018-bottom, lebanoff-etal-2020-cascade, ernst-etal-2021-summary}. 
Conversely, content fusion was largely explored at the full-sentence fusion level \citep{geva-etal-2019-discofuse, lebanoff-etal-2020-learning}, with less emphasis on sub-sentence fusion.




Recently, \citet{slobodkin-etal-2022-controlled, slobodkin2023dont} have proposed a distinct separation of content selection from fusion, treating each as an independent task. They specifically concentrated on fusion, defining it as a standalone task termed \textit{Controlled Text Reduction} (CTR).
This task takes as input pre-selected spans, or `highlights', within an input document, and requires a coherent merging of all the highlighted content, and nothing else.
They also released a designated dataset and several CTR models showing strong adherence to these highlights.

While these studies acknowledged the benefits of decomposing grounded generation to subtasks, they mainly focused on single-document inputs. 
Our work builds on this decomposed approach, extending it to multi-document settings, which introduce new challenges such as managing longer inputs, handling redundant highlights \citep{suzuki-nagata-2017-cutting, calvo2018redundancy}, and dealing with potentially conflicting facts or opinions  \citep{kim2009generating, ma2022multi}.


Additionally, previous CTR studies assessed highlight adherence by comparing outputs with the concatenated highlights, using lexical metrics like ROUGE \citep{lin-2004-rouge} and METEOR \citep{banerjee-lavie-2005-meteor}, and semantic metrics like BERTScore \citep{zhang2020bertscore}.
These methods, while suitable for single-input scenarios, are less effective for multi-document contexts where redundant and conflicting highlights are more prevalent.
Additionally, these approaches did not distinctly evaluate faithfulness and coverage of highlights, typically assessing them jointly, with only manual evaluation for separate evaluation.

Addressing this, we explore more suitable metrics focusing separately on faithfulness and coverage of fused texts, inspired by recent progress in this area.
Several recent studies have used Natural Language Inference (NLI) models for faithfulness evaluation \citep{laban-etal-2022-summac, schuster-etal-2022-stretching}.
There have also been advances in utilizing LLMs to evaluate faithfulness in a zero-shot setting with NLI-style prompts \citep{chen2023evaluating, kocmi-federmann-2023-large, liu2023geval}, or after fine-tuning on synthetic data for faithfulness evaluation \citep{kryscinski-etal-2020-evaluating, yin-etal-2021-docnli, gekhman2023trueteacher}. 
Yet, these works mainly targeted overall source text faithfulness rather than to specific segments.
Moreover, they have not been widely applied to assess coverage, which has traditionally been evaluated using lexical metrics \citep{grusky-etal-2018-newsroom} or manual evaluation \citep{syed-etal-2021-summary}.
In our work, we adapt these methods to our highlights-focused setting, both for faithfulness and coverage, and assess their effectiveness.



 \begin{figure*}[h!]
\centering
    \includegraphics[width=16cm]{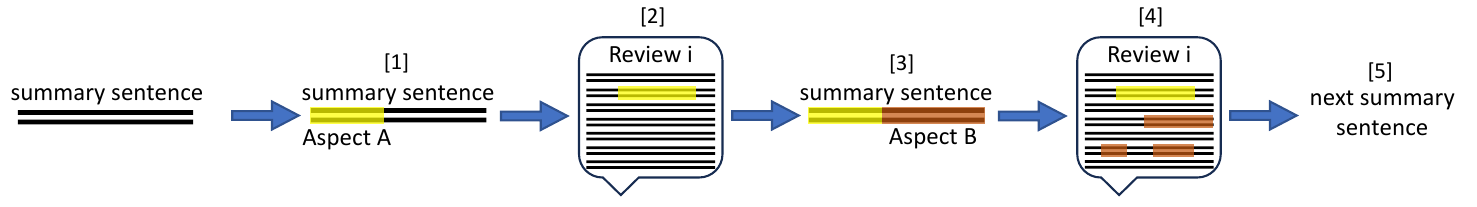}
    \caption{Illustration of the highlighting annotation process for a summary sentence, with reference to a specific review: [1] A \textit{summary} aspect is identified and its statement highlighted; [2] Corresponding \textit{review} spans are highlighted, and the alignment is saved; [3] Another \textit{summary} aspect is identified and highlighted; [4] The matching \textit{review} spans are highlighted, and the alignment is saved; [5] When all summary aspects that are alignable to the current review are highlighted, we proceed to the next sentence, and so on. In this example, the summary consists of two aspects, but steps 1 and 2 can be repeated as needed per sentence, until all alignable aspects are annotated. Borrowed and adapted from \citet{slobodkin-etal-2022-controlled}.}
    \label{fig:annotation_flow}
\end{figure*}

\section{Task Definition}\label{sec:task_definition}
The \textit{Fusion in Context} (\fic{}) task is defined as the process of synthesizing a coherent text from a given set of documents, specifically focusing on pre-selected spans within these documents, referred to as \textit{highlights}.
Formally, given a document set $D$ with marked spans $H=\{h_1, h_2, ..., h_n\}$ (such that $h_i$ may be non-contiguous), a coherent and non-redundant passage $f$ is generated, adhering to the following two criteria: (1) \textit{highlight faithfulness} -- $f$ must be collectively entailed by the content in $H$, adding only minimal non-highlighted content required for coherence; (2) \textit{highlight coverage} -- each $h_i \in H$ must be represented in $f$, either explicitly, or via a generalized reference. For instance, if a highlight states \textit{``the place serves great sushi''}, the output should either directly mention \textit{``great sushi''} or refer to it in more general terms, such as \textit{``great food''}.
Moreover, the task permits the abstraction and aggregation of multiple highlights into a single, synthesized statement. For example, separate highlights noting \textit{``the beds were clean''}, \textit{``the bathrooms were spotless''}, and \textit{``the windows were clean''} could be collectively abstracted to a general statement like \textit{``the rooms are clean''}.
Overall, the goal is to produce a faithful, non-redundant and non-omissive, yet potentially abstractive and aggregated, fusion of the highlighted content.

Next, we describe a dataset (\S{\ref{sec:crowdsourcing}}) and an evaluation framework (\S{\ref{sec:evaluation_framework}}) that comply with the task definition. These are released as the \frev{} benchmark for the \fic{} task.
\section{Dataset for \fic{}}
\label{sec:crowdsourcing}

To comply with the task definition, an instance in a \fic{} dataset is expected to be a document set $D$ with marked spans $H = \{h_1, h_2, ..., h_n\}$, and a corresponding fused text $f$. To compile such data, we leverage existing multi-document summarization datasets and extract high-quality \fic{} instances via controlled crowdsourcing \citep{roit-etal-2020-controlled}, by adapting the method from \citet{slobodkin-etal-2022-controlled} to the multi-text setting, and the business reviews domain. 

\subsection{Dataset Collection}
\label{subsec:Annotation Process}

Given a document set $D$ and corresponding reference summary $\widehat{f}$ from an existing multi-document summarization dataset, the annotation process aims to identify the spans in the source texts $\{h_1, h_2, ..., h_n\}$ that cover all the information within $\widehat{f}$. This approach simplifies the annotation process compared to annotating from scratch, i.e., reading documents, marking highlights according to some specifications, and writing a coherently-fused text, which is reminiscent of standard formation of multi-document summarization datasets. Conversely, our approach requires locating and aligning spans between the source text and the already available reference summary, essentially ``reverse engineering'' the original human summarization process.

\paragraph{Source data.}
For our dataset, we turn to the business reviews domain, and sample review-sets and corresponding summaries from the CocoTrip \citep{iso22aclfindings} and the FewSum \citep{bravzinskas2020few} datasets.
CocoTrip is a dataset of comparative opinion summaries of hotel review-sets, and FewSum consists of summaries of review-sets on businesses. Each review-set in these datasets comprises 8 reviews and up to 6 (average 3.13) corresponding reference summaries. 

\paragraph{Annotation interface.}
To facilitate the annotation of alignments between reviews and their corresponding summary, we adapt a web-based annotation tool from \citet{slobodkin-etal-2022-controlled}, and deploy it on Amazon Mechanical Turk\footnote{\url{www.mturk.com}} for crowdsourcing (\S\ref{subsec:Annotators Training} will explain the controlled crowdsourcing procedure). The application presents reviews and the respective summary side-by-side, and annotators are guided to highlight pairs of spans in the reviews and the summary that directly align. To reduce cognitive load, a summary is displayed alongside an individual review, and focus is placed on one summary sentence at a time. To further ease the process for annotators, lemmas in the review overlapping with lemmas in the currently focused summary sentence are emboldened.\footnote{Lemmatizing with spaCy \cite{spacy2}.} This enables quick skimming through the review, however, workers are trained not to rely solely on exact matches for highlighting (as discussed in \S\ref{subsec:guidelines} and \S\ref{subsec:Annotators Training}).

\paragraph{Annotation procedure.}
Annotators are guided to align statements dealing with a single aspect of a hotel or business (e.g., ``room cleanliness'') from the summary with the most relevant spans in the reviews, and to do so for all summary aspects in order to cover the whole summary text (see \S\ref{subsec:guidelines} for the detailed annotation guidelines).\footnote{We observed that instructing annotators to focus on one aspect at a time enhances the efficiency in locating the relevant review spans, particularly when a summary sentence includes content that is scattered across different parts of the review.}
This, in turn, creates instances of highlighted spans within reviews, with a corresponding coherent fusion of those highlights (the summary).
Each review-summary pair is annotated by a single trained annotator. To enhance quality, submissions are randomly sampled and reviewed, with feedback provided as necessary. 

\begin{table*}[t!]
\centering
\resizebox{\textwidth}{!}{%
\begin{tabular}{lcccccccc}
\hline
        & \textbf{\begin{tabular}[c]{@{}c@{}}\#unique \\ sets \\ of reviews\end{tabular}} & \textbf{\begin{tabular}[c]{@{}c@{}}\#summaries/\\ review-set\\ (average)\end{tabular}} & \textbf{\begin{tabular}[c]{@{}c@{}}\#summary-\\ review-set \\ pairs\end{tabular}} & \textbf{\begin{tabular}[c]{@{}c@{}}mean \\ review/summary \\ size (tokens)\end{tabular}} & \textbf{\begin{tabular}[c]{@{}c@{}}max \\ review/review-set/\\ summary (tokens)\end{tabular}} & \textbf{\begin{tabular}[c]{@{}c@{}}mean \\ review/summary \\ size (sentences)\end{tabular}} & \textbf{\begin{tabular}[c]{@{}c@{}}summary sentences \\ aligning to \\ multiple reviews\end{tabular}} & \textbf{\begin{tabular}[c]{@{}c@{}}summary sentences \\ aligning to multiple \\ review sentences\end{tabular}} \\ \hline
Train   & 237                                                                             & 2.71                                                                                   & 643                                                                               & 87.6/75.18                                                                               & 239/1118/231                                                                                  & 5.89/5.08                                                                                   & 82.51\%                                                                                               & 53.20\%                                                                                                        \\
Dev     & 23                                                                              & 4.30                                                                                   & 99                                                                                & 77.97/69.05                                                                              & 197/829/174                                                                                   & 5.47/4.71                                                                                   & 87.34\%                                                                                               & 57.73\%                                                                                                        \\
Test    & 60                                                                              & 4.30                                                                                   & 258                                                                               & 77.33/67.62                                                                              & 279/881/266                                                                                   & 5.39/4.68                                                                                   & 83.28\%                                                                                               & 51.82\%                                                                                                        \\ \hline
Overall & 320                                                                             & 3.13                                                                                   & 1000                                                                              & 83.99/72.62                                                                              & 279/1118/266                                                                                  & 5.72/4.94                                                                                   & 83.15\%                                                                                               & 53.29\%                                                                                                        \\ \hline
\end{tabular}%
}
\caption{Statistics of our dataset, including the number of unique review-sets, the average number of summaries per review-set, the number of summary/review-set pairs (a unique review-set creates a pair with each of its summaries), the mean review/summary size (in tokens and in sentences), the maximum review/review-set/summary size (in tokens),  the percentage of summary sentences whose alignments span across more than one review, and the percentage of summary sentences whose alignments span across more than one review sentence within one of its reviews (namely, within a single review, the alignments come from more than one sentence).}
\label{tab:dataset_statistics}
\end{table*}

\paragraph{Resulting dataset.}
In total, we sampled 1000 instances of review-set/summary pairs (700 instances from CocoTrip and 300 from FewSum).
See \autoref{tab:dataset_statistics} for full statistics.\footnote{See Appendix~\ref{sec:additional_dataset_details} for more details.}

\subsection{Annotation Guidelines and Data Traits}
\label{subsec:guidelines}
\autoref{fig:annotation_flow} illustrates the annotation flow.
When presented with a review of an entity (hotel or business) and a summary with a sentence in focus, an annotator first identifies aspects of the entity within the focused summary sentence. An aspect is not simply a facet of an entity, such as ``rooms'' or ``staff'', but more specifically it is a characteristic, such as ``room cleanliness'', ``room style'' or ``staff helpfulness'' (more on this in Appendix~\ref{subsec:Summary-related Guidelines}).

Upon identifying an aspect in the summary sentence, annotators are tasked with locating corresponding spans in the review.
These are the minimal spans that adequately cover the information as in the summary regarding the aspect, where omitting any content would miss out on some detail of that aspect in the summary.
For example, omitting any mention of the room being `small' from the review highlights in \autoref{fig:example_task}, would overlook this characteristic of the room, which is mentioned in the second summary sentence.

As outlined in \S\ref{sec:task_definition}, alignments on aspects need to consider two entailment-related traits. Firstly, a summary may express a generalized phrasing of an aspect that is stated in the reviews. For instance, a review may say \textit{``great sushi''} while the summary might just say \textit{``great food''}. Annotators are hence directed to also mark review excerpts that are more specific than in the corresponding spans in the summary. Secondly, several spans in the reviews pertaining to the same aspect may yield an aggregated abstraction in the summary. Annotators must therefore also include review spans that exemplify the summary aspect. For example, aligning a review statement such as \textit{``the beds were clean''} with a summary phrase \textit{``the rooms were clean''}. 

Additionally, reviews often express varying opinions about the same aspect, such as \textit{``the service was great''} as opposed to \textit{``the staff was unprofessional''}. When summarizing, all these varying opinions should be considered to reflect the overall sentiment. As a result, summary segments may range from statements like \textit{``the staff was overall liked''} to \textit{``some people liked the staff''}, depending on the spectrum of opinions.
Hence, to properly capture this consolidation of differing viewpoints, annotators are also guided to align review mentions that either sentimentally entail \textit{or} contradict the summary aspect. For example, the two aforementioned conflicting \textit{review} spans should be aligned to the summary span \textit{``the service was mostly good''}. 

Finally, annotators may mark multiple spans in reviews that redundantly represent the same statement. The guidelines also address paraphrasing, non-consecutive highlights, and unalignable summary spans. A detailed explanation of these guidelines can be found in Appendix~\ref{subsec:Review-related Guidelines}.

\subsection{Annotator Training}
\label{subsec:Annotators Training}

The requirements of the aforementioned annotation process call for proficient-level annotations, which we achieved by means of controlled crowdsourcing \cite{roit-etal-2020-controlled}. We identified qualified annotators through three open qualification rounds, followed by three closed rounds for selected annotators, focusing on further training and refinement.
Each open round involved annotators reading a brief task description and accordingly aligning information between a single summary sentence and a short review, on a simplified interface. After each open round, we reviewed the alignment and provided feedback. We then checked whether the annotators implemented our feedback in the following round (with a different sentence-review instance). If the annotators satisfyingly cooperated throughout the open rounds, they moved on to the closed rounds. Before the closed rounds, the qualified workers were asked to watch a 25-minute tutorial on the full annotation tool and guidelines (\S\ref{subsec:Annotation Process} and \S\ref{subsec:guidelines}).
The closed rounds were conducted similarly to the open rounds, but with a whole summary and review, with all guidelines, and on the full interface. The qualification process was fully compensated with a customary wage, requiring up to 5 minutes per round.
From this process, we were able to gather 8 trained annotators, who annotated the 1000 instances in our dataset.

\subsection{Dataset Quality}\label{subsec: Dataset Quality}
To evaluate the quality of the compiled dataset, we compute the inter-annotator agreement.
To this end, for every two annotators, we calculate intersection-over-union (\textit{IoU}) of the tokens' indices (considering only content words) between the highlighted review spans that are aligned to the same summary sentence, similarly to \citet{ernst-etal-2021-summary}.
The \textit{IoU} scores are gathered on the sentence level across three review-set/summary pairs, annotated by six crowdworkers.
The resulting IoU score is 61.8.

To better understand the sources of disagreements, we analyzed all cases when $IoU<90\%$. 
We found that the main cause of disagreement was related to our criteria for generalization and aggregation. Here, some annotators chose specific review spans they believed exemplified a summary characteristic, while others opted for different spans. This does not harm the quality of our data, as in all cases, the summary segment was indeed aligned with each of the corresponding review spans, according to our criteria.
Another common source of disagreement involved annotators including additional phrases that provided only insignificant extra details on top of the summary.
For detailed examples, refer to Appendix~\ref{sec:IAA_disagreement_Examples}.

Finally, an interesting aspect of our dataset is that 80\% of the summary sentences align to spans from multiple reviews, and over $50\%$ of the summary sentences align with non-consecutive spans from different sentences within a single review (see \autoref{tab:dataset_statistics}). We also find that on average, 25\% of a review's tokens were highlighted. These properties reflect the real-world challenges faced by \fic{} models, expected to coherently fuse disparate, and at times redundant, details.

\section{Evaluation Framework}
\label{sec:evaluation_framework}

Consistent with the task definition in \S\ref{sec:task_definition}, a passage produced by a model as a fusion of highlights within source documents must uphold several criteria. (1) \textit{Faithfulness}: it must only contain information from the highlights; (2) \textit{Coverage}: it must cover all the information in the highlights, be it in an explicit, generalized, or aggregated form; (3) \textit{Coherence and Redundancy}: it must convey the information in a well-structured and non-redundant form.
In this section, we suggest several automatic metrics for faithfulness and coverage, and assess their effectiveness by correlating to human scores that we collected. Coherence and redundancy are measured using manual evaluation.

\subsection{Limitations of Lexical and Semantic Matching}
\label{subsec:lexical_and_semantic_metrics}
Output's adherence to highlights was previously measured in \citet{slobodkin-etal-2022-controlled, slobodkin2023dont} 
by comparing the output passage and the concatenated highlights, using lexical metrics like \textbf{ROUGE} (n-gram matching) and \textbf{METEOR} (word matching with synonyms), and semantic metrics like  \textbf{BERTScore} (probability of generating the output text). 
Our work, however, extends beyond the single-document scenario explored in these previous works, to also include multi-document contexts.
This shift introduces additional complexities, such as managing redundancy and contradictions among highlights drawn from diverse sources, which may not be fully captured by standard lexical and semantic matching techniques.
Further, our setting also enables highlights aggregation and generalization, which these metrics may not adequately address.
Additionally, these automated approaches primarily measured overall adherence to the highlights without making a distinction between faithfulness and coverage. These latter aspects were evaluated manually, but only on a limited number of instances.


\subsection{NLI-based Faithfulness Metric}\label{subsec:nli_faithfulness_metrics}
Highlight-faithfulness requires the output passage to be entailed by the collective highlighted content. We employ the \texttt{flan-t5-xxl} model \citep{https://doi.org/10.48550/arxiv.2210.11416}, shown to exhibit high performance on NLI tasks, for evaluating faithfulness to highlights in a zero-shot setting with a standard \textbf{NLI} prompt (see Appendix~\ref{sec:nli_zero_shot_prompt}).
Previous research that used NLI models for faithfulness evaluation in summarization \citep{maynez-etal-2020-faithfulness, laban-etal-2022-summac, honovich-etal-2022-true-evaluating} typically set the grounding text as the premise, and the generated text as the hypothesis. Accordingly, we set the highlights concatenation to serve as the premise, since the outputs are expected to be entailed by all the highlighted content collectively (see \S\ref{sec:task_definition}).
For the hypothesis, we segment the output passage into sentences, with each sentence serving as a separate hypothesis.
The average of the sentence-level entailment scores is used as the overall entailment probability of the corresponding passage. This approach, inspired by \citep{laban-etal-2022-summac}, was found to be more effective than using the entire output as a single hypothesis.\footnote{We also experimented with other methods for evaluating faithfulness and coverage, which exhibited lower correlation to human judgment. See Appendix~\ref{sec:additional_evaluation_framework_details} for more details.}

\subsection{Trained Coverage Metric}\label{subsec:trained_coverage_metric}
Inspired by recent work that evaluates faithfulness and factuality using a dedicated \textbf{trained} model \citep{yin-etal-2021-docnli, utama-etal-2022-falsesum, gekhman2023trueteacher, soleimani-etal-2023-nonfacts}, we finetune an LLM that is tasked to assess whether the generated passage fully covers the highlights. 
In our methodology, each highlight is individually input along with the entire output, and the model outputs a binary answer for whether the highlight is contained in the passage.\footnote{We also tried concatenating all the highlights together, and found it to be inferior.}
We derive synthesized training data for this task from our \fic{} dataset, using highlights and their corresponding summaries. For negative samples, we remove the summary sentence that aligns with the highlight. For positive samples, a random non-aligning summary sentence is omitted (to avoid a potential bias caused by sentence exclusion in the negative samples).
We finetune a \texttt{flan-t5-large} model \citep{https://doi.org/10.48550/arxiv.2210.11416} with the synthesized coverage data.
The input to the model is the highlight and modified summary, and the output is `yes' or `no', for positive and negative samples, respectively. The final score is the average probability of the token `yes' across all highlights.\footnote{We also explored an NLI-based coverage metric, where the passage serves as the premise and the highlights function as the hypothesis. We found it to achieve comparable results, however it requires substantially more computation time and memory. For more details, see Appendix~\ref{sec:additional_evaluation_framework_details}.}

\begin{table}[]
\centering
\resizebox{\columnwidth}{!}{%
\begin{tabular}{l|cc|cc}
\hline
\textbf{}                   & \multicolumn{2}{c|}{{\ul \textbf{Faithfulness}}} & \multicolumn{2}{c}{{\ul \textbf{Coverage}}}     \\
\textbf{Metric}             & \textbf{$\tau$}  & \textbf{{\small CI}} & \textbf{ $\tau$} & \textbf{{\small CI}} \\ \hline
{\small ROUGE-1 (R)}        & 0.2319                    & 0.23-0.24            & 0.3467                   & 0.34-0.35            \\ \hline
{\small ROUGE-1 (P)}        & 0.5468                    & 0.54-0.55            & -0.0533                  & -0.06--0.05          \\ \hline
{\small ROUGE-2 (R)}        & 0.3555                    & 0.35-0.36            & 0.2731                   & 0.27-0.28            \\ \hline
{\small ROUGE-2 (P)}        & 0.5253                    & 0.52-0.53            & 0.0071                   & 0.00-0.01            \\ \hline
{\small ROUGE-L (R)}        & 0.0958                    & 0.09-0.10            & 0.3835                   & 0.38-0.39            \\ \hline
{\small ROUGE-L (P)}        & 0.4898                    & 0.48-0.49            & -0.0367                  & -0.04--0.03          \\ \hline
{\small METEOR}             & 0.4017                    & 0.40-0.41            & 0.2736                   & 0.27-0.28            \\ \hline
{\small BERTScore (R)}      & 0.2380                    & 0.23-0.24            & 0.4165                   & 0.41-0.42            \\ \hline
{\small BERTScore (P)}      & 0.6004                    & 0.59-0.60            & 0.0529                   & 0.05-0.06            \\
\midrule \midrule
{\small NLI (Faithfulness)} & \textbf{0.6745}           & \textbf{0.67-0.68}   & 0.0929                   & 0.09-0.10            \\ \hline
{\small Trained (Coverage)} & 0.1771                    & 0.17-0.18            & \textbf{0.4992}          & \textbf{0.49-0.50}   \\ \hline
\end{tabular}%
}
\caption{Average Kendall-Tau rank correlations ($\tau$) and their $95\%$ confidence intervals (CI) for tested evaluation metrics against human judgment. Recall-based metrics (R) are more effective for coverage, and precision-based metrics (P) for faithfulness. Best correlations for each axis are in bold. \os{maybe we don't need the F1-based metrics in the table?} \os{the recall and precision is not explained in the text}}
\label{tab:correlation_with_human_judgement}
\end{table}

\subsection{Meta-Evaluation}\label{subsec:meta_evaluation}

\paragraph{Setup.}
To assess our evaluation metrics we follow the common practice \citep{fabbri2021summeval} of correlating scores to human judgment. To that end, we gather faithfulness and coverage ratings for generated outputs from three co-authors of this paper.
The outputs were produced by two models (see Flan-T5\textsubscript{H} and Flan-T5\textsubscript{no-H} in \S\ref{subsec:experimental_setup}).
A total of 50 review sets were randomly selected from our test set, leading to 100 scores for each of coverage and faithfulness.
A 1-to-7 Likert scale was used to rate faithfulness and coverage separately for an output.

To ensure agreement among annotators, the three authors first evaluated a separate set of 10 outputs, and inter-annotator agreement was computed with Cohen's Kappa coefficient \citep{cohen1960}.
The average Kappa coefficients were 0.49 and 0.42 for faithfulness and coverage, respectively, indicating a moderate level of agreement \citep{viera2005understanding}.
For more details, see Appendix~\ref{sec:additional_meta_evaluation_setup_details}.

After collecting scores for the 100 instances, we computed their correlation with human judgment using Kendall-Tau rank correlation, as suggested in \citep{deutsch2022re}.\footnote{Spearman correlations were also calculated, showing similar trends. See Appendix~\ref{sec:additional_meta_evaluation_results}.}
We also apply bootstrapping \citep{efron1987better} by performing 1000 samplings of 70 instances (with repetition) and calculating correlation scores for each such subset. We report the average correlation and 95\% confidence intervals for each metric.

\paragraph{Results.} 
\autoref{tab:correlation_with_human_judgement} shows the average correlations with their $95\%$ confidence intervals for faithfulness and coverage. We find that while certain lexical- and semantic-based metrics yield decent results, notably BERTScore-precision for faithfulness and BERTScore-recall for coverage, our proposed metrics demonstrate significantly higher correlations, with average values of 0.6745 and 0.4992 for faithfulness and coverage, respectively.
In light of these findings, we employ our NLI-based and trained metrics for assessing model performance in terms of faithfulness and coverage, respectively (in \S\ref{subsec:results}).

\subsection{Human Evaluation of Coherence and Redundancy}
We adopt the coherence assessment methodology from \citep{slobodkin-etal-2022-controlled}. Crowdworkers judge the coherence of 100 randomly selected instances from the test set, for each examined model. A score between 1 and 5 is specified, and each passage is reviewed by three workers and averaged. Similarly, the redundancy of information in a passage is appraised.
This approach follows standard practice, where coherence and redundancy are best evaluated manually \citep{fabbri2021summeval, steen-markert-2021-evaluate}. 
For more details see Appendix~\ref{sec:Fluency_and_Reduancy_Human_Annotation_Protocol}.
\begin{table*}[th]
\centering
\resizebox{0.7\textwidth}{!}{%
\begin{tabular}{l|ccc|cc}
\toprule
\textbf{Model}                & \textbf{Faithfulness} & \textbf{Coverage} & \textbf{F-1} & \textbf{Coherence} & \textbf{Redundancy} \\ \midrule
Flan-T5\textsubscript{H}      & 72.8                  & 86.4              & 79.0         & 4.3                & 4.1                 \\
Flan-T5\textsubscript{H} (RL) & 54.0                  & 82.0              & 65.1         & 4.1                & 4.0                 \\
Flan-T5\textsubscript{only-H} & \textbf{84.6}                  & \textbf{87.8}              & \textbf{86.2}         & 3.6                & 3.8                 \\
Flan-T5\textsubscript{no-H}   & 53.7                  & 76.9              & 63.2         & 4.1                & 3.9                 \\
GPT-4                         & 81.6                  & 85.6              & 83.6         & \textbf{4.7}                & \textbf{4.5}                 \\ \bottomrule
\end{tabular}%
}
\caption{Results for the proposed models on our \fic{} dataset. Faithfulness is measured with our NLI-based metric, and Coverage with our trained metric. The F-1 is a harmonic mean of the two latter scores. Coherence and Redundancy are measured through manual assessment. For each metric, the best score is in bold. \ran{If you rank each of the four Flan methods according to each of the four metrics (ignoring F1) Then T5H has the highest average rank across all four. Maybe add this info because it shows you need both documents and highlights to succeed.}}
\label{tab:results}
\end{table*}
\section{Experiments}\label{sec:experiments}

\subsection{Experimental Setup}\label{subsec:experimental_setup}

We examine several baseline models for solving the \fic{} task. The input to a model is a document set with spans marked within the documents (highlights), and the model is trained to generate a fused passage around the highlights.

\paragraph{Models with full input.}
Using the training set of our dataset, we finetune a large language model, marking the highlights in the input via designated mark-ups, following \citet{slobodkin-etal-2022-controlled}. Specifically, we finetune a \texttt{flan-t5-large} model \citep{https://doi.org/10.48550/arxiv.2210.11416}, that exhibited enhanced performance in tasks requiring constrained generation \citep{sanh2022multitask, wei2022finetuned}. We will refer to this model as \textbf{Flan-T5\textsubscript{H}} (`H' for `Highlights').
We develop an additional variant of Flan-T5\textsubscript{H}, which we further finetune using Reinforcement Learning (\textbf{RL}), following the method in \citet{slobodkin2023dont}. It applies the Quark algorithm \citep{lu2022quark} combined with a dual-reward policy \citep{pasunuru-bansal-2018-multi}, alternating between our NLI-based faithfulness and trained coverage metrics (\S\ref{sec:evaluation_framework}) as rewards.
We also examine the performance of a one-shot \textbf{GPT-4} model \citep{openai2023gpt4}, guided with an example of the task.\footnote{Preliminary experiments on a separate development set, with varying numbers of in-context examples, indicated that a single exemplar yields the best results. See Appendix~\ref{sec:gpt4_prompting}.}

\paragraph{Models with highlights only.}
To reveal the importance of the surrounding context, we also train a \texttt{flan-t5-large} model only with a concatenation of the highlights as the input (excluding surrounding context). We denote this variant \textbf{Flan-T5\textsubscript{only-H}}.

\paragraph{Models without highlights.}
Finally, we examine \texttt{flan-t5-large} in a standard summarization setting, where it is finetuned with the input review-set without the highlighted spans, denoting this variant \textbf{Flan-T5\textsubscript{no-H}}. It offers insights into the model's ability to pick up on signals that point to highlights.


\subsection{Results}\label{subsec:results}
We apply our evaluation metrics on the proposed systems, with results presented in \autoref{tab:results}. We first observe that the exclusion of context from the input (Flan-T5\textsubscript{only-H}) yields the strongest faithfulness and coverage scores, yet the lowest coherence and redundancy scores. This shows the importance of incorporating context for more seamless outputs. Meanwhile, the removal of highlights (Flan-T5\textsubscript{no-H}) leads to a substantial degradation in faithfulness and coverage. This indicates that the Flan-T5\textsubscript{H} model indeed succeeds in learning to adjust the output according to the highlights, underlining the highlights' role in enhancing the model's performance.

Interestingly, even though the RL reward functions used in the RL-enriched model are the faithfulness and coverage metrics themselves, the outputs are eventually negatively affected when evaluating with these metrics\os{do we have an intuition for why?}. This result calls for a more in-depth investigation of enhanced reward functions that can leverage the benefits of RL training, as was shown to be helpful in \citet{slobodkin2023dont} for the single-input setup. We also find that single-shot GPT-4 yields the most coherent and least redundant texts. While it ranks highly in faithfulness and coverage, it is still overtaken by the finetuned Flan-T5\textsubscript{only-H}. Overall, our findings invite for further research on the \fic{} task, to develop fusion strategies that ensure comprehensive coverage and faithfulness to highlighted content, with coherent and low-redundancy outputs.

\section{Conclusion}\label{sec:conclusion}
In this paper, we further promote the decomposition of grounded text generation as presented in \citep{slobodkin-etal-2022-controlled}, extending it to the multi-document setting.
To that end, we introduce the Fusion-in-Context (\fic{}) task, an extension of the task from \citep{slobodkin-etal-2022-controlled} which focuses on the content fusion step, to the multi-document setting.
The \fic{} setting facilitates employing a single general-purpose fusion model for diverse content selection needs, capturing the challenges of repetitiveness and contradictions in source documents. 
It also supports interactive, user-driven generation, by allowing users to choose personalized content for the \fic{} module to merge into a customized passage.
Moreover, direct access to the pre-selected ``highlights'' can facilitate attributed generation, where the pre-selected segments also serve as supporting cited content for the fused text.
To advance the task, we introduce the \frev{} benchmark, which includes a high-quality dataset, an evaluation framework for faithfulness and coverage of selected spans, and several baseline models to stimulate further research and exploration.

Future work may include expanding the \fic{} task to other multi-input contexts, e.g., the news domain.
We also plan to investigate ways to leverage the built-in traceability of the output text's origin, namely the highlights, for facilitating attributed generation.
\section{Limitations}\label{sec:limitations}
In this work, we construct the first \fic{} dataset, developed by instructing crowdworkers to identify relevant spans within reviews that align with the content of corresponding summaries.
To reduce cognitive load, each summary was displayed alongside individual reviews. 
While this approach streamlined the annotation process, there are instances where viewing the complete set of input reviews is advantageous, particularly for aggregative summary segments.
In such segments, multiple review spans are combined into a single summary span, necessitating a broader understanding of the entire input set for accurate highlighting.

Moreover, the focus of our dataset on the business reviews domain may constrain its generalizability to other contexts with distinct textual structures, like news articles.
This limitation extends to our trained evaluation metrics, which were developed using a derivative of our crowdsourced dataset and, therefore, are tailored to the specific characteristics of business reviews.

\section{Ethics Statement}\label{sec:ethics}
The proposed Fusion-in-Context (\fic{}) task, despite offering enhanced control over the content generated, is not expected to achieve complete resolution. Therefore, integrating \fic{} modules in modular generative systems should be done so with caution, since there is a possibility that these modules may overlook certain highlighted content or inadvertently include content that was not highlighted. 
This concern is particularly relevant for future endeavors that aim to use \fic{} for attributed generation. In such cases, there is a risk that some portions of the generated content may not be directly traceable to the pre-defined highlighted segments, leading to potential inaccuracies, or incompleteness, in attribution.


\bibliography{anthology,custom}

\appendix

\section{Annotation Full Guidelines}\label{sec:Annotation Full Guidelines}
In this section, we provide the full annotation guidelines, presented to our workers.

\subsection{Summary-related Guidelines}\label{subsec:Summary-related Guidelines}
As mentioned in \S\ref{subsec:guidelines}, we guide annotators to segment summary sentences into the different aspects of hotels or businesses.
The annotation guidelines distinguish between two classifications of aspects: \\
$\bullet$ \textsc{different aspects}: This refers to independent facets of the business, e.g., location and room quality. \\
$\bullet$ \textsc{different characteristics of the same aspect}: This pertains to addressing varied characteristics within the same aspect, for example, the cleanliness and size of a room.

\subsection{Review-related Guidelines}\label{subsec:Review-related Guidelines}
This section provides a detailed overview of the review-related guidelines presented to our crowdworkers during their training: \\
$\bullet$ \textsc{any mention of the aspect}: Annotators are trained to align all review mentions of a summary aspect, encompassing both similar and contrasting sentiments. For instance, if the summary aspect is \textit{``The staff was friendly''}, both positive and negative mentions regarding staff friendliness are to be aligned. \\
$\bullet$ \textsc{specificity in reviews}: 
Crowdworkers are advised to align review mentions that offer more specificity than the summary aspects. For example, a general summary statement like \textit{``The staff was helpful''}, should be aligned with a more specific review comment, such as \textit{``the concierge was very helpful''}. We also emphasize that the other way around, namely, that the summary is more specific than the reviews, should not be aligned. \\ 
$\bullet$ \textsc{exemplification in reviews}: In line with the previous point, annotators are guided to focus on identifying review segments that provide examples of the summary statements.  An example would be aligning the summary span \textit{``The hotel is well-maintained''} with a review segment that exemplifies it, such as \textit{``the pool area is very clean''}. As in the previous point, we discourage our crowdworkers from considering the reverse cases, when the summaries exemplify the reviews. \\
$\bullet$ \textsc{paraphrasing}: Annotators are instructed to align paraphrased mentions in reviews with the summary content, such as aligning \textit{``the hotel is overpriced''} with \textit{``you can stay at lovely B\&B in the old town that is actually cheaper than this''}. \\
$\bullet$ \textsc{consecutiveness}: We guide our workers to avoid highlighting unnecessary details, i.e., that did not appear in the summary span, and keep the highlights inconsecutive if needed. \\
$\bullet$ \textsc{unalignable spans}: Recognizing that each summary is derived from multiple reviews, but reviewers assess only one review at a time, it is often the case that not all summary details will be present in the reviewed content. In such instances, annotators are instructed to leave such summary spans unhighlighted.

\begin{figure*}[t]

\lstdefinestyle{promptStyle}
{
    basicstyle={\footnotesize\ttfamily},
    numbers=left,numberstyle=\footnotesize,
    xleftmargin=2.8em,
    xrightmargin=1.5em,
    showstringspaces=false,
      showspaces=false,
        showtabs=false,
    tabsize=2,
    breaklines=true,
        flexiblecolumns=true,
        escapeinside={<@}{@>},
          breakatwhitespace=true
}

\newtcblisting{mylisting}[1]{
  enhanced,
  listing only,
  boxrule=0.8pt,
  sharp corners=downhill,
  top=0mm,
  bottom=0mm,
  left=2mm,
  right=0mm,
  boxsep=0mm,
  colframe=black,
  colback=white,
  listing options={
    style=#1
  }
}

\definecolor{instructionsColor}{rgb}{0.1, 0.5, 0.1}

\begin{mylisting}{promptStyle}
<@\textcolor{instructionsColor}{\#\#\# Instruction: Read the following and determine if the hypothesis can be inferred from the premise.}@>
<@\textcolor{instructionsColor}{Options: Entailment, Contradiction, or Neutral}@>

<@\textcolor{red}{\#\#\# Input:}@>
<@\color{blue}Premise:@>  {Premise}
<@\color{blue}Hypothesis:@> {Hypothesis}

<@\color{red} \#\#\# Response (choose only one of the options from above):@> 

\end{mylisting}
\caption{The prompt structure employed in zero-shot configurations as a basis for evaluating the frameworks of faithfulness and coverage.}
\label{fig:nli_zero_shot_prompt}
\end{figure*}

\section{NLI Zero-Shot Prompt}\label{sec:nli_zero_shot_prompt}
\autoref{fig:nli_zero_shot_prompt} demonstrates the structure of the zero-shot prompt used for the nli-based evaluation frameworks of highlights coverage and faithfulness.

\begin{table}[t!]
\centering
\resizebox{\columnwidth}{!}{%
\begin{tabular}{cccc}
\hline
Number of Exemplars & Faithfulness  & Coverage      & F-1           \\ \hline
1                   & \textbf{80.1} & 85.0          & \textbf{82.5} \\
2                   & 72.1          & \textbf{86.1} & 78.5          \\
3                   & 73.6          & 84.0          & 78.5          \\
4                   & 72.8          & 82.2          & 77.2          \\ \hline
\end{tabular}%
}
\caption{Faithfulness, coverage, and F-1 scores of the zero-shot GPT-4 model on 30 instances for the \fic{} development set, for varying numbers of in-context examples in the prompt. For each metric, the best scores are in bold.}
\label{tab:GPT-4_icl_examples_tuning}
\end{table}

\section{GPT-4 Prompting}\label{sec:gpt4_prompting}
\autoref{tab:GPT-4_icl_examples_tuning} presents the faithfulness, coverage, and F-1 scores of the zero-shot GPT-4 model across 30 instances from the \fic{} development set, for varying numbers of in-context examples in the prompt. Based on these outcomes, we chose to proceed with a single in-context example.

\section{Fluency and Redundancy Human Annotation Protocol}\label{sec:Fluency_and_Reduancy_Human_Annotation_Protocol}
\begin{figure*}[t!]
    \centering
    \setcounter{subfig}{0} 

    \begin{subfigure}{\linewidth}
        \centering
        \includegraphics[width=1\linewidth]{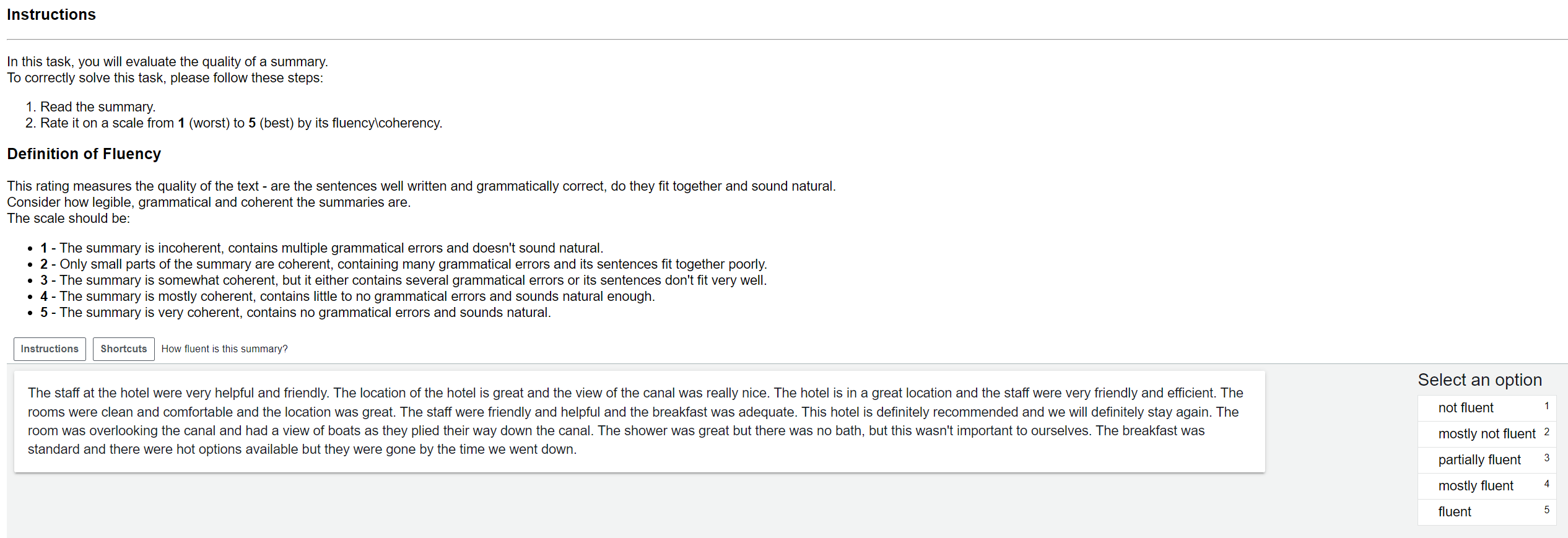}
        \stepcounter{subfig} 
        \caption{(\alph{subfig}) Fluency Evaluation Interface}
        \label{subfig:coherency_interface} 
    \end{subfigure}
    
    \begin{subfigure}{\linewidth}
        \centering
        \includegraphics[width=1\linewidth]{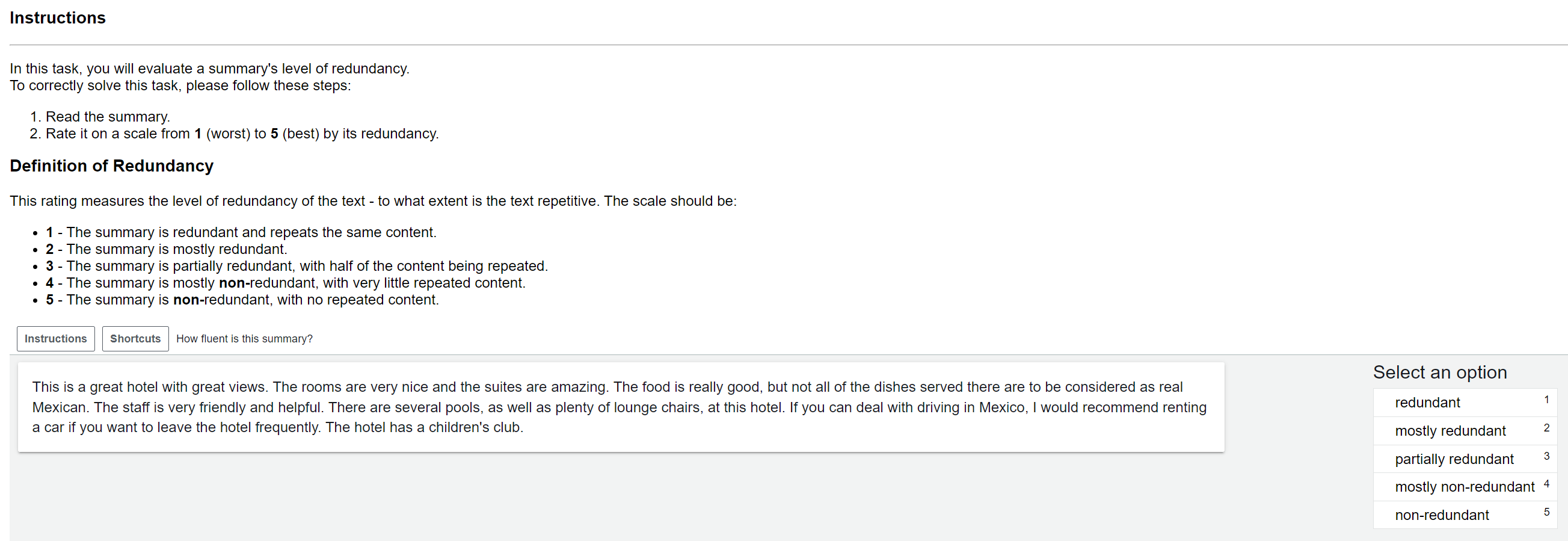}
        \stepcounter{subfig} 
        \caption{(\alph{subfig}) Redundancy Evaluation Interface}
        \label{subfig:redundancy_interface} 
    \end{subfigure}

    \caption{Example of the data collection interfaces used by the crowd-workers to evaluate the fluency (\ref{subfig:coherency_interface}) and redundancy (\ref{subfig:redundancy_interface}) of summaries.}
    \label{fig:coherency_and_redundancy_interfaces} 
\end{figure*}

We ask crowd-workers to assess the fluency and redundancy of the texts produced by all models under examination.
We employ annotators who have demonstrated proficiency in semantic tasks, including summarization, in previous experiments.
For evaluation purposes, 100 instances are randomly selected from our test set, and the texts generated by each model for these instances are evaluated, resulting in 500 total samples. Each sample is reviewed by three different annotators, and their scores are averaged to obtain a final assessment.
The evaluation is facilitated through two Amazon Mechanical Turk interfaces, specifically designed for this study. One interface focuses on evaluating coherence, while the other assesses redundancy, with each interface presenting the annotators with one of the 500 samples (as depicted in \autoref{fig:coherency_and_redundancy_interfaces}).
Consistent with the methodology of \citep{slobodkin-etal-2022-controlled}, a 5-point Likert scale is employed to rate the fluency and redundancy of the generated summaries. To minimize ambiguity and promote consistent ratings, each score on the scale is accompanied by explicit criteria (also illustrated in \autoref{fig:coherency_and_redundancy_interfaces}).
Taking into account an average response time of 30 seconds for each evaluation, we set the compensation for each response at 10 \textcentoldstyle.

\section{Additional Evaluation Framework Details}\label{sec:additional_evaluation_framework_details}

\subsection{Trained Faithfulness Metric}\label{subsec:trained_faithfulness_metric}
In a similar fashion to the trained coverage metric, we
we use our crowdsourced dataset to generate training data for evaluating highlights faithfulness. This approach mirrors the NLI-based metric we proposed, wherein a model is trained to individually evaluate the faithfulness of each output sentence, subsequently averaging the scores across all sentences. 

For the positive training instances, we separate each summary from our crowdsourced dataset into sentences, and pair each sentence with all the instance's highlights. In contrast, for the creation of negative instances, we remove all highlights that were aligned with any segment of the corresponding summary sentence.
The training process involves fine-tuning a \texttt{flan-t5-large} model \citep{https://doi.org/10.48550/arxiv.2210.11416}. 
In this setup, the input comprises the highlights and the summary sentence, while the output is either the token 'yes' for positive instances, or 'no' for negative ones.  The final score is calculated based on the probability assigned to the token 'yes' by the model.

\subsection{NLI-based Coverage Metric}\label{subsec:nli_based_coverage_metric}
For the evaluation of highlight-coverage using Natural Language Inference (NLI), our approach mirrors the one implemented for assessing faithfulness using NLI (see \S\ref{subsec:nli_faithfulness_metrics}), albeit with a role reversal, where the output serves as the premise and the highlights function as the hypothesis. Rather than treating all highlights collectively as the hypothesis, we calculate the coverage of each highlight separately and then average across all highlights.\footnote{We consider each individual alignment in our crowdsourced dataset as a distinct highlight.}

\begin{table}[t]
\centering
\resizebox{0.6\columnwidth}{!}{%
\begin{tabular}{ccc}
\hline
Judges & Faithfulness & Coverage \\ \hline
1-2    & 0.37         & 0.31     \\
2-3    & 0.71         & 0.67     \\
1-3    & 0.39         & 0.27     \\ \hline
\end{tabular}%
}
\caption{The individual Cohen's Kappa coefficients for each pair of judges, on the faithfulness and coverage axes.}
\label{tab:cohen_kappa_individual_coefficients}
\end{table}

\subsection{Additional Meta Evaluation Setup Details}\label{sec:additional_meta_evaluation_setup_details}
\paragraph{Pairwise Cohen Kappa Coefficients}
\autoref{tab:cohen_kappa_individual_coefficients} shows the pairwise Cohen's Kappa coefficients for each pair of judges.

\paragraph{Reconciliation Process}
To achieve further agreement between the three authors, an additional reconciliation procedure was undertaken for the ten instances annotated by all three authors. 
This procedure entailed discussions for each instance where the annotations diverged by more than one point, separately for the faithfulness and coverage scores.
During these discussions, each author explained the rationale behind their assigned score. Subsequently, the authors endeavored to reach a unanimous agreement on each instance, thereby further aligning their scoring criteria.



\begin{table}[ht!]
\centering
\resizebox{\columnwidth}{!}{%
\begin{tabular}{l|cc|cc}
\hline
\textbf{}                       & \multicolumn{2}{c|}{{\ul \textbf{Faithfulness}}} & \multicolumn{2}{c}{{\ul \textbf{Coverage}}}     \\
\textbf{Metric}                 & \textbf{{\small $\tau$}}  & \textbf{{\small CI}} & \textbf{{\small $\tau$}} & \textbf{{\small CI}} \\ \hline
{\small ROUGE-1 (R)}            & 0.2319                    & 0.23-0.24            & 0.3467                   & 0.34-0.35            \\ \hline
{\small ROUGE-1 (P)}            & 0.5468                    & 0.54-0.55            & -0.0533                  & -0.06--0.05          \\ \hline
{\small ROUGE-1 (F1)}           & 0.5587                    & 0.55-0.56            & 0.1497                   & 0.14-0.16            \\ \hline
{\small ROUGE-2 (R)}            & 0.3555                    & 0.35-0.36            & 0.2731                   & 0.27-0.28            \\ \hline
{\small ROUGE-2 (P)}            & 0.5253                    & 0.52-0.53            & 0.0071                   & 0.00-0.01            \\ \hline
{\small ROUGE-2 (F1)}           & 0.4964                    & 0.49-0.50            & 0.1477                   & 0.14-0.15            \\ \hline
{\small ROUGE-L (R)}            & 0.0958                    & 0.09-0.10            & 0.3835                   & 0.38-0.39            \\ \hline
{\small ROUGE-L (P)}            & 0.4898                    & 0.48-0.49            & -0.0367                  & -0.04--0.03          \\ \hline
{\small ROUGE-L (F1)}           & 0.3880                    & 0.38-0.39            & 0.1950                   & 0.19-0.20            \\ \hline
{\small METEOR}                 & 0.4017                    & 0.40-0.41            & 0.2736                   & 0.27-0.28            \\ \hline
{\small BERTScore (R)}          & 0.2380                    & 0.23-0.24            & 0.4165                   & 0.41-0.42            \\ \hline
{\small BERTScore (P)}          & 0.6004                    & 0.59-0.60            & 0.0529                   & 0.05-0.06            \\ \hline
{\small BERTScore (F1)}         & 0.4958                    & 0.49-0.50            & 0.2555                   & 0.25-0.26            \\ \midrule \midrule
{\small NLI (Faithfulness)}     & \textbf{0.6745}           & \textbf{0.67-0.68}   & 0.0929                   & 0.09-0.10            \\ \hline
{\small NLI (Coverage)}         & 0.2255                    & 0.22-0.23            & \textbf{0.5084}          & \textbf{0.50-0.51}   \\ \hline
{\small Trained (Faithfulness)} & 0.5836                    & 0.58-0.59            & 0.2495                   & 0.24-0.25            \\ \hline
{\small Trained (Coverage)}     & 0.1771                    & 0.17-0.18            & 0.4992                   & 0.49-0.50            \\ \hline
\end{tabular}%
}
\caption{Average Kendall-Tau rank correlations ($\tau$) and their $95\%$ confidence intervals (CI) for tested evaluation metrics against human judgment. Recall-based metrics (R) are more effective for coverage, and precision-based metrics (P) for faithfulness. Best correlations for each axis are in bold.}
\label{tab:full_correlation_with_human_judgement_kendall}
\end{table}

\begin{table}[h!]
\centering
\resizebox{\columnwidth}{!}{%
\begin{tabular}{l|cc|cc}
\hline
\textbf{}                       & \multicolumn{2}{c|}{{\ul \textbf{Faithfulness}}} & \multicolumn{2}{c}{{\ul \textbf{Coverage}}}     \\
\textbf{Metric}                 & \textbf{{\small $\tau$}}  & \textbf{{\small CI}} & \textbf{{\small $\tau$}} & \textbf{{\small CI}} \\ \hline
{\small ROUGE-1 (R)}            & 0.3124                    & 0.31-0.32            & 0.4440                   & 0.44-0.45           \\ \hline
{\small ROUGE-1 (P)}            & 0.6892                    & 0.68-0.69            & -0.0861                  & -0.09--0.08          \\ \hline
{\small ROUGE-1 (F1)}           & 0.7172                    & 0.71-0.72            & 0.1654                   & 0.16-0.17            \\ \hline
{\small ROUGE-2 (R)}            & 0.4842                    & 0.48-0.49            & 0.3537                   & 0.35-0.36            \\ \hline
{\small ROUGE-2 (P)}            & 0.6807                    & 0.68-0.69            & 0.0005                   & -0.01-0.01           \\ \hline
{\small ROUGE-2 (F1)}           & 0.6590                    & 0.65-0.66            & 0.1885                   & 0.18-0.20            \\ \hline
{\small ROUGE-L (R)}            & 0.1237                    & 0.12-0.13            & 0.4902                   & 0.48-0.50            \\ \hline
{\small ROUGE-L (P)}            & 0.6420                    & 0.64-0.65            & -0.0596                  & -0.07--0.05          \\ \hline
{\small ROUGE-L (F1)}           & 0.5160                    & 0.51-0.52            & 0.2531                   & 0.25-0.26            \\ \hline
{\small METEOR}                 & 0.5412                    & 0.54-0.55            & 0.3487                   & 0.34-0.36            \\ \hline
{\small BERTScore (R)}          & 0.3141                    & 0.31-0.32            & 0.5237                   & 0.52-0.53            \\ \hline
{\small BERTScore (P)}          & 0.7450                    & 0.74-0.75            & 0.0485                   & 0.04-0.06            \\ \hline
{\small BERTScore (F1)}         & 0.6516                    & 0.65-0.66            & 0.3267                   & 0.32-0.33            \\ \midrule \midrule
{\small NLI (Faithfulness)}     & \textbf{0.8257}           & \textbf{0.82-0.83}   & 0.1088                   & 0.10-0.12            \\ \hline
{\small NLI (Coverage)}         & 0.2831                    & 0.28-0.29            & \textbf{0.6355}          & \textbf{0.63-0.64}   \\ \hline
{\small Trained (Faithfulness)} & 0.7268                    & 0.72-0.73            & 0.3271                   & 0.32-0.33            \\ \hline
{\small Trained (Coverage)}     & 0.2315                    & 0.22-0.24            & 0.6178                   & 0.61-0.62            \\ \hline
\end{tabular}%
}
\caption{Average Spearman's rank correlations ($\tau$) and their $95\%$ confidence intervals (CI) for tested evaluation metrics against human judgment. Recall-based metrics (R) are more effective for coverage, and precision-based metrics (P) for faithfulness. Best correlations for each axis are in bold.}
\label{tab:full_correlation_with_human_judgement_spearman}
\end{table}

\subsection{Additional Meta Evaluation Results}\label{sec:additional_meta_evaluation_results}
Tables~\ref{tab:full_correlation_with_human_judgement_kendall} and \ref{tab:full_correlation_with_human_judgement_spearman} present the full correlations with human judgments using the Kendall-Tau rank correlations and Spearman's rank correlations, respectively, including the additional evaluation frameworks we explored (see Appendices~\ref{subsec:trained_faithfulness_metric} and \ref{subsec:nli_based_coverage_metric}), and the F-1 scores for the ROUGE and BERTScore metrics.

\begin{table*}[t]
\centering
\resizebox{\textwidth}{!}{%
\begin{tabular}{lcccccccc}
\toprule
         & \textbf{\begin{tabular}[c]{@{}c@{}}\vspace{-1.5mm} {\small\#unique} \\ \vspace{-1.5mm} {\small sets} \\ {\small of reviews}\end{tabular}} & \textbf{\begin{tabular}[c]{@{}c@{}}\vspace{-1.5mm}{\small\#summaries/}\\ \vspace{-1.5mm}{\small review-set}\\ {\small(average)}\end{tabular}} & \textbf{\begin{tabular}[c]{@{}c@{}}\vspace{-1.5mm}{\small \#summary-}\\ \vspace{-1.5mm}{\small review-set} \\ {\small pairs}\end{tabular}} & \textbf{\begin{tabular}[c]{@{}c@{}}\vspace{-1.5mm} {\small mean} \\ \vspace{-1.5mm}{\small review/summary} \\ {\small size (tkns)}\end{tabular}} & \textbf{\begin{tabular}[c]{@{}c@{}} \vspace{-1.5mm}{\small max} \\ \vspace{-1.5mm}{\small review/review-set/}\\ {\small summary (tkns)}\end{tabular}} & \textbf{\begin{tabular}[c]{@{}c@{}} \vspace{-1.5mm}{\small mean} \\ \vspace{-1.5mm}{\small review/summary} \\ {\small size (sents)}\end{tabular}} & \textbf{\begin{tabular}[c]{@{}c@{}} \vspace{-1.5mm}{\small summary sents} \\ \vspace{-1.5mm}{\small aligning to} \\ {\small multiple reviews}\end{tabular}} & \textbf{\begin{tabular}[c]{@{}c@{}}\vspace{-1.5mm}{\small summary sents} \\  \vspace{-1.5mm}{\small aligning to multiple} \\ {\small review sents} \end{tabular}} \\
\midrule
Train    & \multicolumn{1}{l}{}                                                         & \multicolumn{1}{l}{}                                                                   & \multicolumn{1}{l}{}                                                              & \multicolumn{1}{l}{}                                                                   & \multicolumn{1}{l}{}                                                                        & \multicolumn{1}{l}{}                                                                    & \multicolumn{1}{l}{}                                                                                  & \multicolumn{1}{l}{}                                                                                           \\
\hspace{2mm} {\small CocoTrip} & 184                                                                          & 2.63                                                                                   & 484                                                                               & 97.56/80.15                                                                            & 239/1118/231                                                                                & 6.22/5.32                                                                               & 80.74\%                                                                                               & 50.02\%                                                                                                         \\
\hspace{2mm} {\small FewSum}   & 53                                                                           & 3.00                                                                                   & 159                                                                               & 57.27/60.06                                                                            & 75/497/104                                                                                  & 4.87/4.34                                                                               & 89.13\%                                                                                               & 65.07\%                                                                                                         \\
\hspace{2mm} {\small Total}      & 237                                                                          & 2.71                                                                                   & 643                                                                               & 87.6/75.18                                                                             & 239/1118/231                                                                                & 5.89/5.08                                                                               & 82.51\%                                                                                               & 53.20\%                                                                                                         \\
Dev      & \multicolumn{1}{l}{}                                                         & \multicolumn{1}{l}{}                                                                   & \multicolumn{1}{l}{}                                                              & \multicolumn{1}{l}{}                                                                   & \multicolumn{1}{l}{}                                                                        & \multicolumn{1}{l}{}                                                                    & \multicolumn{1}{l}{}                                                                                  & \multicolumn{1}{l}{}                                                                                           \\
\hspace{2mm} {\small CocoTrip} & 10                                                                           & 6.00                                                                                   & 60                                                                                & 91.24/75.93                                                                            & 197/829/174                                                                                 & 5.64/5.07                                                                               & 85.53\%                                                                                               & 46.05\%                                                                                                         \\
\hspace{2mm} {\small FewSum}   & 13                                                                           & 3.00                                                                                   & 39                                                                                & 57.55/58.46                                                                            & 78/493/102                                                                                  & 5.21/4.15                                                                               & 90.74\%                                                                                               & 79.63\%                                                                                                         \\
\hspace{2mm} {\small Total}      & 23                                                                           & 4.30                                                                                   & 99                                                                                & 77.97/69.05                                                                            & 197/829/174                                                                                 & 5.47/4.71                                                                               & 87.34\%                                                                                               & 57.73\%                                                                                                         \\
Test     & \multicolumn{1}{l}{}                                                         & \multicolumn{1}{l}{}                                                                   & \multicolumn{1}{l}{}                                                              & \multicolumn{1}{l}{}                                                                   & \multicolumn{1}{l}{}                                                                        & \multicolumn{1}{l}{}                                                                    & \multicolumn{1}{l}{}                                                                                  & \multicolumn{1}{l}{}                                                                                           \\
\hspace{2mm} {\small CocoTrip} & 26                                                                           & 6.00                                                                                   & 156                                                                               & 90.34/74.28                                                                            & 279/881/266                                                                                 & 5.64/4.86                                                                               & 79.95\%                                                                                               & 42.74\%                                                                                                         \\
\hspace{2mm} {\small FewSum}   & 34                                                                           & 3.00                                                                                   & 102                                                                               & 57.43/57.44                                                                            & 74/509/105                                                                                  & 5.0/4.41                                                                                & 88.89\%                                                                                               & 67.11\%                                                                                                         \\
\hspace{2mm} {\small Total}    & 60                                                                           & 4.30                                                                                   & 258                                                                               & 77.33/67.62                                                                            & 279/881/266                                                                                 & 5.39/4.68                                                                               & 83.28\%                                                                                               & 51.82\%                                                                                                         \\ 
\midrule
Overall     & \multicolumn{1}{l}{}                                                         & \multicolumn{1}{l}{}                                                                   & \multicolumn{1}{l}{}                                                              & \multicolumn{1}{l}{}                                                                   & \multicolumn{1}{l}{}                                                                        & \multicolumn{1}{l}{}                                                                    & \multicolumn{1}{l}{}                                                                                  & \multicolumn{1}{l}{}                                                                                           \\
\hspace{2mm} {\small CocoTrip}  & 220                                                                          & 3.18                                                                                   & 700                                                                               & 95.41/78.48                                                                            & 279/1118/266                                                                                & 6.04/5.2                                                                                & 80.97\%                                                                                               & 48.17\%                                                                                                         \\
\hspace{2mm} {\small FewSum}     & 100                                                                          & 3.00                                                                                   & 300                                                                               & 57.36/58.96                                                                            & 78/509/105                                                                                  & 4.96/4.34                                                                               & 89.25\%                                                                                               & 67.59\%                                                                                                         \\
\hspace{2mm} {\small Total}      & 320                                                                          & 3.13                                                                                   & 1000                                                                              & 83.99/72.62                                                                            & 279/1118/266                                                                                & 5.72/4.94                                                                               & 83.15\%                                                                                               & 53.29\%                                                                                                         \\ 
\bottomrule
\end{tabular}%
}
\caption{Full statistics of our dataset, including the number of unique review-sets, the average number of summaries per review-set, the number of summary/review-set pairs (a unique review-set creates a pair with each of its summaries), the mean review/summary size (in tokens and in sentences), the maximum review/review-set/summary size (in tokens),  the percentage of summary sentences whose alignments span across more than one review, and the percentage of summary sentences whose alignments span across more than one review sentence within one of its reviews (namely, within a single review, the alignments come from more than one sentence).}
\label{tab:full_dataset_statistics}
\end{table*}



\section{Additional Dataset Details}\label{sec:additional_dataset_details}

\subsection{Full \fic{} Dataset Statistics}\label{subsec:full_FiC_dataset_Statistic}
\autoref{tab:full_dataset_statistics} presents the full \fic{} dataset statistics, including specific statistics for each of the dataset's splits and instances origin, i.e., CocoTrip or FewSum.

\subsection{Annotation Cost}\label{subsec:annotation_cost}
Each annotation instance, averaging 4 minutes, is priced at 70\textcentoldstyle. 
Annotators also receive compensation for training activities, including a 5$\$$ bonus for taking the 25-minute tutorial and an additional 2$\$$ for reviewing feedback. The total cost for the dataset amounted to approximately 5700$\$$.

\subsection{Additional Details about the Annotators Recruitment}
For our crowdsourcing project, we hired annotators from English-speaking countries who had over 5000 approved HITs as well as an approval rate higher than 98\% on Amazon Mechanical Turk. During the recruitment process, in addition to explaining the annotation guidelines, we also explained to the crowdworkers the purpose of the dataset, in order to rationalize different aspects of the annotation protocol.

\begin{figure*}[t!]
    \centering
    \begin{subfigure}[h]{1.0\textwidth}
       \includegraphics[width=1\linewidth]{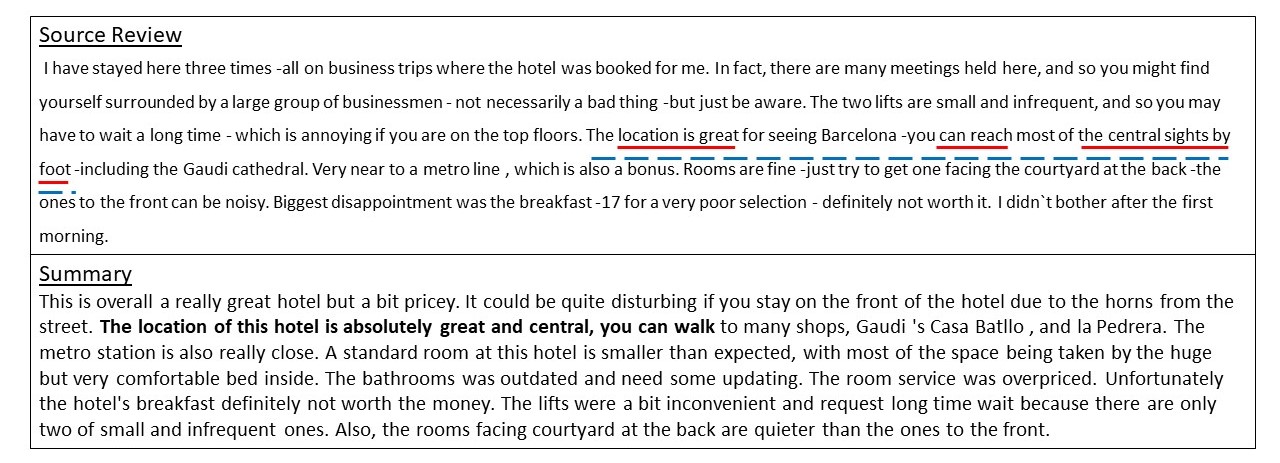}
       \caption{}
       \label{fig:IAA_disagreement_example1} 
    \end{subfigure}
    
    \begin{subfigure}[h]{1.0\textwidth}
       \includegraphics[width=1\linewidth]{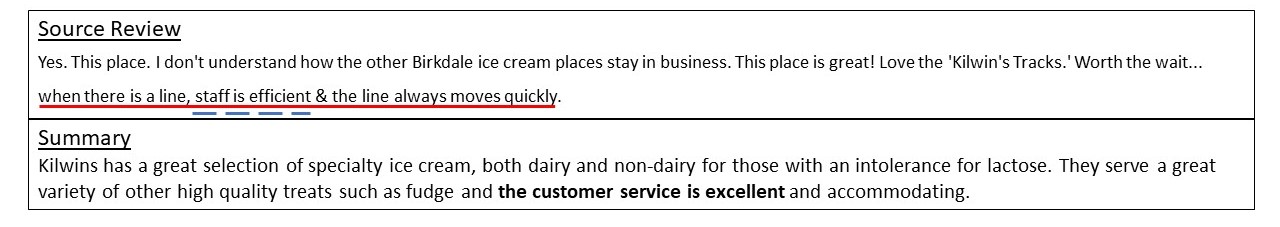}
       \caption{}
       \label{fig:IAA_disagreement_example2}
    \end{subfigure}
    
    \caption{Two examples of disagreement between annotators. For each example, the bottom part is the summary (with the summary span over which there was disagreement in bold) and the top part is a review with both the annotators' highlights (marked with a red solid line and a blue dashed line to indicate each highlight).}
    \label{fig:IAA_disagreement_examples}
\end{figure*}

\section{IAA disagreement Examples}\label{sec:IAA_disagreement_Examples}
\autoref{fig:IAA_disagreement_examples} demonstrates two instances of disagreements between our annotators.

\section{Additional Experimental Details}\label{sec:appendix_experimental_details}
To incorporate the highlighting signal in the baseline Flan-T5\textsubscript{H}, \textit{<extra\_token\_1>} and \textit{<extra\_token\_2>} tokens were added to the input, before and after each highlight.
For all trained models, we set the maximum input length to 2048, to accommodate the input length of the language model. 

We also set the maximum target length to 200, which we found works best, as well as setting the batch size to 1.
The other parameters are similar to \citet{slobodkin-etal-2022-controlled, slobodkin2023dont}.
The model is trained for 10k steps.
Training is performed on two A100-SXM4-80GB GPUs, and costs about 12 GPU hours for the supervised models (Flan-T5\textsubscript{H}, Flan-T5\textsubscript{no-H}, and Flan-T5\textsubscript{only-H}) and about 36 GPU hours for the RL-tuned variant of Flan-T5\textsubscript{H}.

Additionally, to train the trained faithfulness and coverage evaluators, we concatenate the highlights concatenation and the output's sentence (for faithfulness) and the generated output with each of the highlights (for coverage), and use the special token \textit{<extra\_token\_4>} as a delimiter.  
For both evaluators, we set the maximum input length to 1024, the maximum target length to 4
, and the batch size to 1.
We train the models for 10 epochs.
Training is performed on a single A100-SXM4-80GB GPU, and costs about 4 GPU hours.

Overall, our trained models, both for faithfulness and coverage evaluation and for the \fic{} task, use \texttt{flan-t5-large} as their backbone model, which consists of 780 million parameters, and our zero-shot NLI-based evaluation frameworks use \texttt{flan-t5-xxl} as the backbone model, consisting of 11 billion parameters.

 \begin{figure*}[h!]
\centering
    \includegraphics[width=16cm]{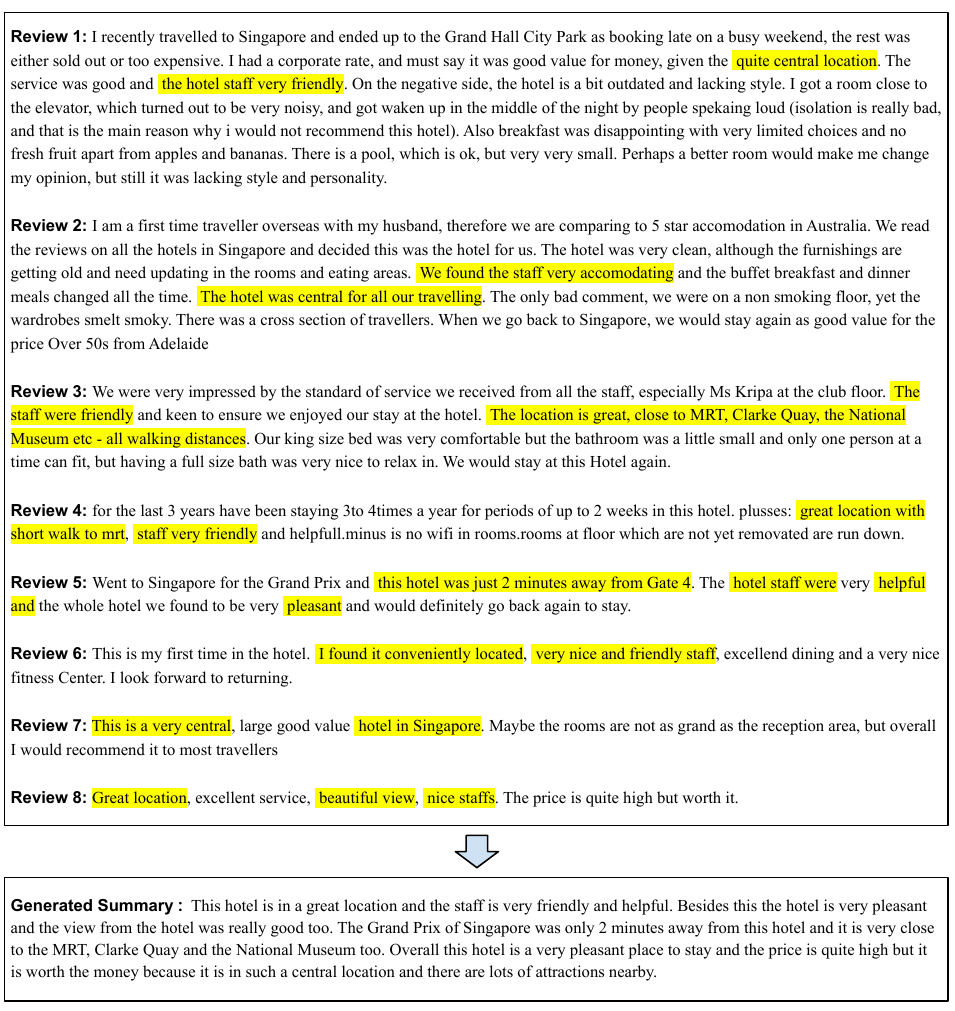}
    \caption{Example of generated output by Flan-T5\textsubscript{H}.}
    \label{fig:generation_example1}
\end{figure*}

 \begin{figure*}[h!]
\centering
    \includegraphics[width=16cm]{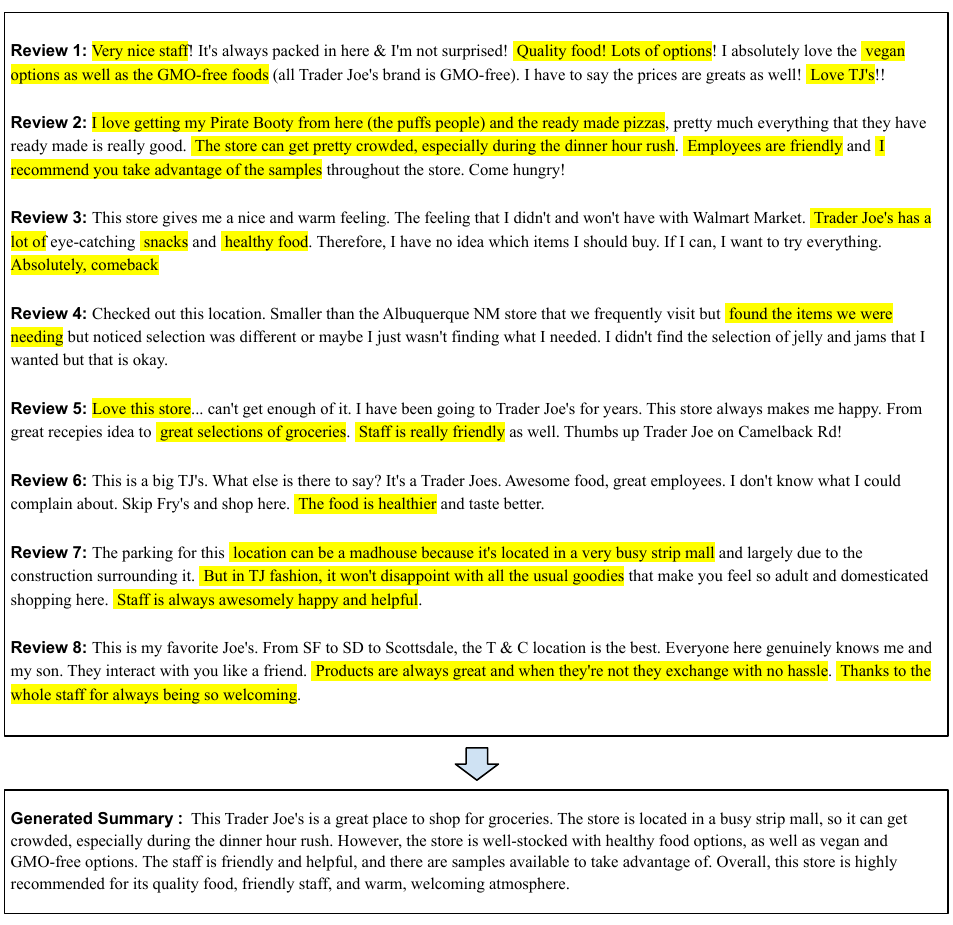}
    \caption{Example of generated output by Flan-T5\textsubscript{H}.}
    \label{fig:generation_example2}
\end{figure*}

\section{Generation Examples}\label{sec:generation_examples}
\autoref{fig:generation_example1} and \autoref{fig:generation_example2} demonstrate two examples of highlighted reviews and the corresponding output generated by Flan-T5\textsubscript{H}.

\begin{table}[]
\centering
\resizebox{0.8\columnwidth}{!}{%
\begin{tabular}{lccc}
\hline
\textbf{Model}               & \textbf{Faithfulness} & \textbf{Coverage} & \textbf{F-1} \\ \hline
\texttt{flan-t5-small} & 66.9                  & 83.4              & 74.2         \\
\texttt{flan-t5-base}  & 70.1                  & 85.5              & 77.0         \\
\texttt{flan-t5-large} & 72.8                  & 86.4              & 79.0         \\ \hline
\end{tabular}%
}
\caption{Faithfulness, coverage, and F-1 results on the \fic{} testset, for different sizes of \texttt{flan-t5} (all finetuned on the \fic{} trainset).}
\label{tab:Flan-T5_sizes}
\end{table}

\section{Impact of Model Size}\label{sec:impact_of_model_size}
\autoref{tab:Flan-T5_sizes} illustrates the performance on the test set of \texttt{flan-t5} with different model sizes, finetuned on the \fic{} trainset.

\section{List of Data and Software Licenses Employed in this Paper}
Our framework  dependencies are:
\begin{enumerate}
    \item CocoTrip dataset: \url{https://github.com/megagonlabs/cocosum/blob/main/LICENSE}, under an Apache License 2.0.
    \item FewSum dataset: \url{https://github.com/abrazinskas/FewSum/blob/master/LICENSE.txt}, under the MIT License.
    \item Quark: \url{https://github.com/GXimingLu/Quark}, Misc.
    \item Baseline model for the zero-shot NLI-based evaluation frameworks: \url{https://huggingface.co/google/flan-t5-xxl/tree/main}, under an Apache License 2.0.
    \item Baseline model for the trained evaluation frameworks and models: \url{https://huggingface.co/google/flan-t5-large/tree/main}, under an Apache License 2.0.
\end{enumerate}

\end{document}